\documentclass[letterpaper, 10pt, conference]{ieeeconf}  

\IEEEoverridecommandlockouts                              

\usepackage{cite}
\usepackage{multicol}
\usepackage[bookmarks=true]{hyperref}
\usepackage{xcolor}
\usepackage{soul}
\usepackage{graphicx}
\usepackage{subcaption}
\usepackage{cuted}
\usepackage{copyright} 
\usepackage{array}
\newcolumntype{P}[1]{>{\centering\arraybackslash}p{#1}}  
\newcolumntype{M}[1]{>{\centering\arraybackslash}m{#1}}  
\usepackage{booktabs}
\usepackage{mathtools}

\usepackage{csquotes}
\usepackage{amsmath}
\usepackage{amssymb}
\usepackage{adjustbox}

\newcommand{\robotdemosubfigwidth}{0.49\columnwidth}

\newcommand{\architectureacronym}{COBRA-PPM}

\newcommand{\predacc}{88.6\%}
\newcommand{\fscore}{0.909}
\newcommand{\precision}{0.955}
\newcommand{\recall}{0.868}
\newcommand{\aucscore}{0.961}

\newcommand{\ourmethodtaskperformance}{94.2\%}

\newcommand{\taskattemptserealrobot}{5}
\newcommand{\tasksuccessesrealrobot}{4}

\begin{document}

\title{\LARGE \bf
\architectureacronym: A Causal Bayesian Reasoning Architecture Using Probabilistic Programming for Robot Manipulation Under Uncertainty}

\author{
Ricardo Cannizzaro$^{1}$,
Michael Groom$^{1}$,
Jonathan Routley$^{1}$, 
Robert Ness$^{2}$, 
and Lars Kunze$^{1,3}$
\thanks{$^{1}$Oxford Robotics Institute, Dept. Engineering Science, University of Oxford, UK. Correspondence email: ricardo@robots.ox.ac.uk}
\thanks{$^{2}$Microsoft Research, Redmond, WA, USA. $^{3}$Bristol Robotics Laboratory, School of Engineering, University of the West of England, Bristol, UK}
\thanks{This work is supported by the Australian Defence Science \& Technology Group, Dyson Technology, and the EPSRC RAILS project (EP/W011344/1).}%
}

\IEEEaftertitletext{
    \begin{center}
        \includegraphics[width=0.95\textwidth, trim=0cm 0cm 0cm 0cm, clip]{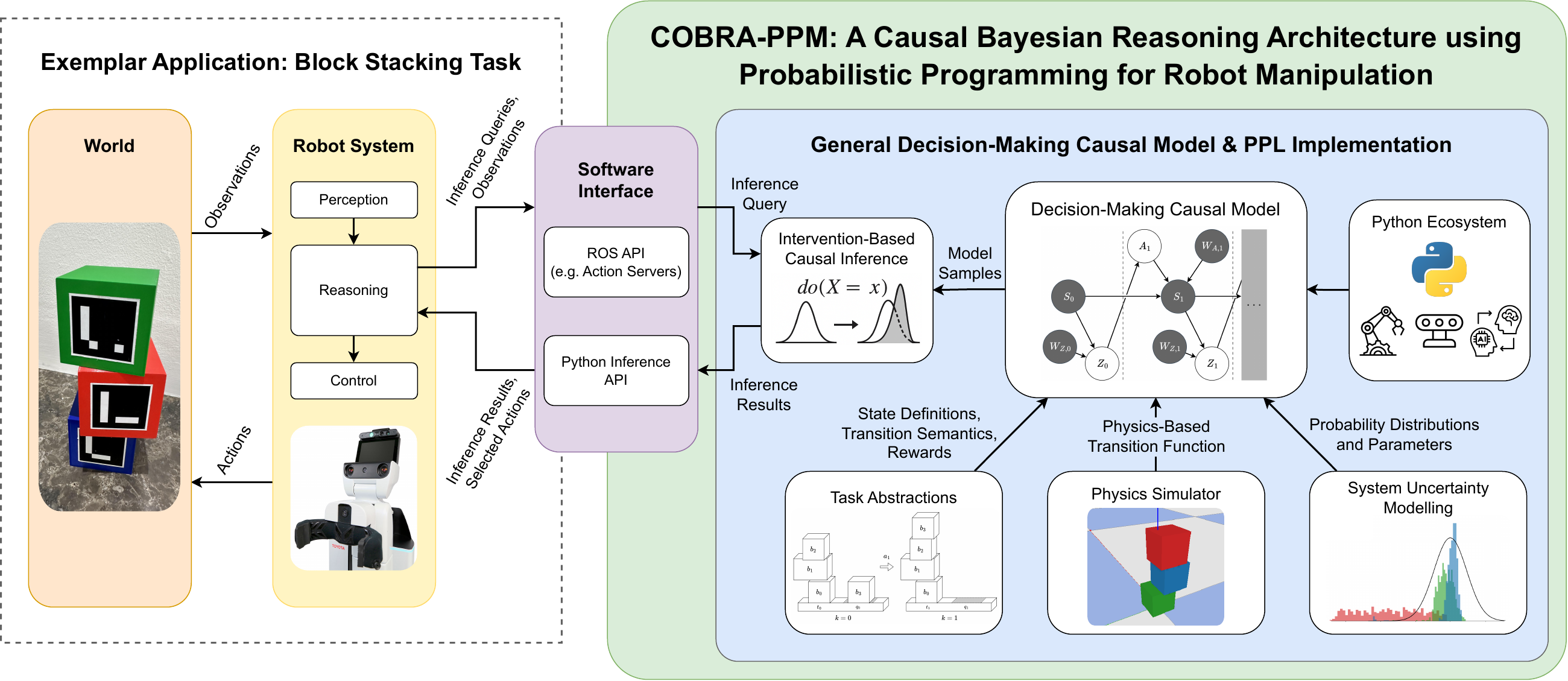}
        \captionof{figure}{
        \architectureacronym: our proposed causal reasoning architecture for robot manipulation under uncertainty, combining \textbf{causal Bayesian networks} and \textbf{probabilistic programming}, and integrated with a full robotic system for an exemplar block-stacking task.
        The figure illustrates both \textbf{generalisable, task-agnostic components} (green box) and \textbf{application-specific components}. 
        The generalisable components include a Pyro-based causal model with causal inference methods, integration with the Python ecosystem, and Python/ROS interfaces enabling seamless deployment across diverse robot platforms. 
        The application-specific components demonstrate how the architecture is instantiated for the block-stacking task, including the world environment, robot hardware, and the \textbf{flexible composition of CBN subcomponents}: task abstractions, the PyBullet physics simulator, and robot system modelling modules.
        }
        \label{fig:reasoning_framework_architecture_diagram}
    \end{center}
}

\maketitle

\copyrightnotice 

\begin{abstract}
Manipulation tasks require robots to reason about cause and effect when interacting with objects. Yet, many data-driven approaches lack causal semantics and thus only consider correlations.
We introduce \emph{\architectureacronym}, a novel causal Bayesian reasoning architecture that combines causal Bayesian networks and probabilistic programming to perform interventional inference for robot manipulation under uncertainty.
We demonstrate its capabilities through high-fidelity Gazebo-based experiments on an exemplar block stacking task, where it: (1) predicts manipulation outcomes with high accuracy (Pred Acc: \predacc); and (2) performs greedy next-best action selection with a \ourmethodtaskperformance\ task success rate. 
%
We further demonstrate sim2real transfer on a domestic robot, showing effectiveness in handling real-world uncertainty from sensor noise and stochastic actions. Our generalised and extensible framework supports a wide range of manipulation scenarios and lays a foundation for future work at the intersection of robotics and causality.
%
\end{abstract}

\vspace{-2mm}
\section{Introduction}\label{sec:introduction}
%
Robot manipulation is central to applications such as warehouse logistics and domestic service robotics, and remains a long-standing focus of research~\cite{suomalainen2022robotcontactsurvey,collins2023ramp,luo2021robust}.
Despite substantial progress, purely data-driven approaches often fail in novel scenarios not seen during training, which can lead to unsafe execution. Real-world robots also face multiple sources of uncertainty~\cite{Kurniawati2022POMDPReview} \textemdash\ such as partial and noisy observations and stochastic actions \textemdash\ which further challenge generalisation.

Model-based methods offer a principled alternative by incorporating knowledge of system dynamics, enabling reasoning in unforeseen circumstances. In manipulation tasks, this requires understanding the probabilistic causal relationships governing object interactions. However, real-world robot dynamics are often non-linear and complex, making faithful simulation and data generation difficult.

While deep learning has shown promise for learning manipulation skills, such methods typically operate at the level of statistical association, thus lack mechanisms for causal reasoning (i.e., interventions). 
They do not model symbolic or causal structure \textemdash\ both essential for building explainable and trustworthy autonomous systems.
Probabilistic programming languages (PPLs) enable the implementation of generative models as programs, offering a flexible and principled foundation for causal reasoning in robotics. Their structured and modular design supports generalisable model construction and reasoning under uncertainty.

We propose \emph{\architectureacronym}: a novel causal Bayesian reasoning architecture using probabilistic programming for robot manipulation under uncertainty. 
For the first time, our approach combines causal Bayesian network (CBN) modelling and inference, probabilistic programming, and online physics simulation to support generalisable, extensible robot decision-making in uncertain real-world environments. 
Crucially, our architecture explicitly accounts for perception and actuation noise, delivering robust performance across sequential placements despite cumulative uncertainty.
We validate the proposed architecture in high-fidelity simulation and on a real domestic robot, demonstrating its predictive accuracy and robustness in block stacking tasks under uncertainty.
Our main contributions are:
\begin{itemize}
    \item The formulation of \architectureacronym, a novel causal reasoning architecture using probabilistic programming for robot manipulation under uncertainty (\textbf{Sec.~\ref{sec:architecture}});
    \item The design of software interfaces (Python and ROS) that enable practical integration of the architecture into real and simulated robot systems (\textbf{Sec.~\ref{subsec:software_interface_for_hardware_integration}});
    \item A real-world hardware demonstration of our architecture using a domestic service robot (\textbf{Sec.~\ref{sec:robot_demonstration}}).
\end{itemize}
\begin{figure}[t]
    \centering
    \includegraphics[trim=1cm 0cm 0cm 0cm, clip,width=1.0\columnwidth]{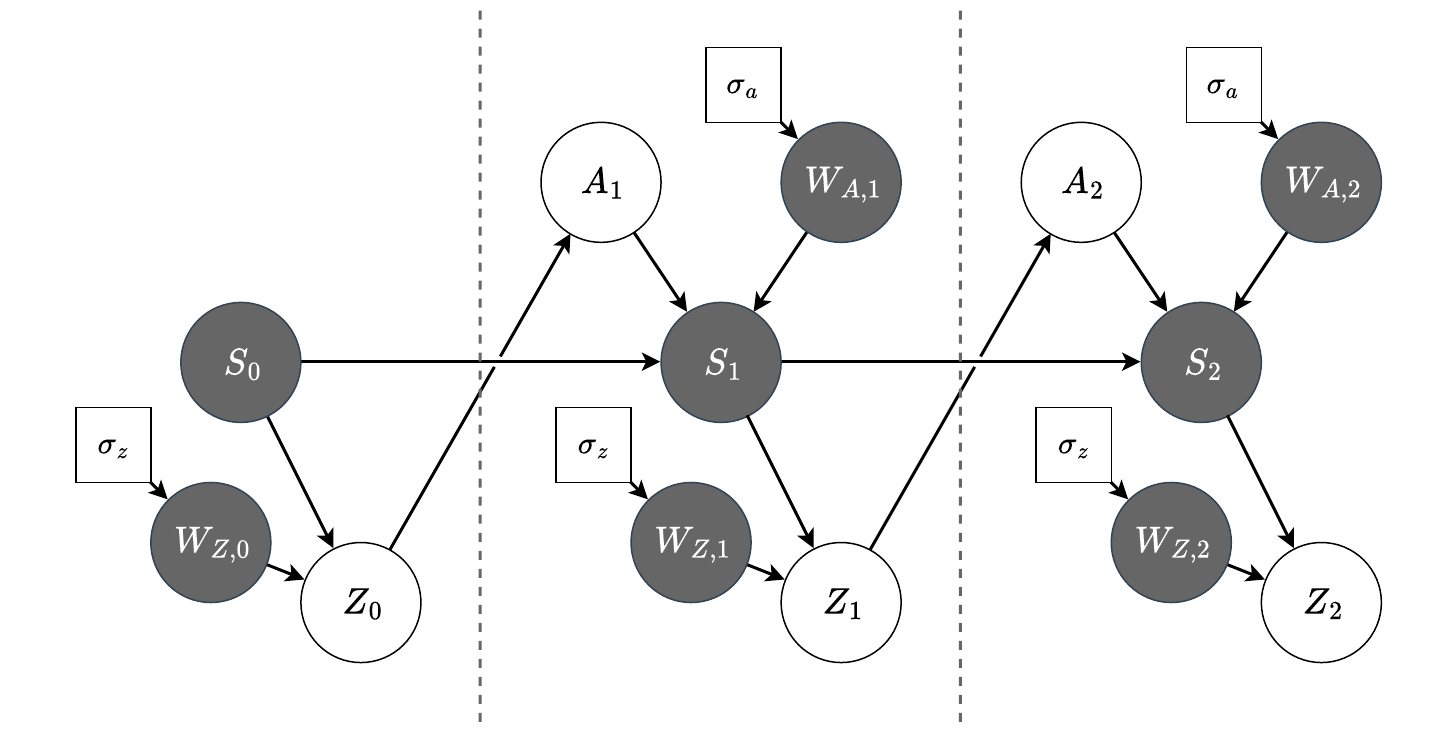}
    \caption{A dynamic causal Bayesian network (CBN) used as the decision-making model within our reasoning architecture (Fig.~\ref{fig:reasoning_framework_architecture_diagram}), illustrated for a two-action block-stacking task (Sec.~\ref{sec:exemplar_block_stacking_task}). The CBN encodes a POMDP structure tailored to the task, capturing domain-specific causal dependencies and probabilistic uncertainty.
    At each time step $k$, the robot selects a stochastic action $A_k$ based on a noisy observation $Z_{k-1}$ of the previous state. 
    The action is executed with noise $W_{A,k}$, leading to a transition from the latent state $S_{k-1}$ to $S_k$, followed by a new noisy observation $Z_k$ subject to $W_{Z,k}$.
    Observed variables are shown as white nodes, unobserved variables as grey nodes, and fixed model parameters as squares.
    }
    \label{fig:block_stacking_task_dag}
    \vspace{-5mm}
\end{figure}
\section{Related Work}\label{sec:related_work}
Causality has gained increased attention as a foundation for trustworthy robot autonomy~\cite{Hellstrom2021CausationInRobotics,Lake2017BuildingMachines}, addressing key limitations of machine learning such as domain transfer and lack of interpretability or accountability~\cite{Pearl2019CausalInferenceAndML,Ganguly2023}. 
Recent work has applied causality to diverse robotics domains: manipulation, navigation, and autonomous vehicles~\cite{Cannizzaro2023CARDESPOT,Howard2023CounterfactualCausalDiscoveryOnDriverBehaviour}.

In robot manipulation, causal reasoning remains underexplored. Diehl \& Ramirez-Amaro~\cite{Diehl2022WhyDidIFail} use a CBN to model block stacking, performing action selection using outcome probabilities learned from offline simulations. However, their model lacks configurability, ignores sensor and actuator uncertainty, and must be retrained for new tasks. Follow-up work~\cite{Diehl2023CausalExplainPredictPreventFailures} addresses failure prediction, but inherits the same limitations. In contrast, our method performs online simulation at inference time, enabling dynamic reconfiguration and explicit modelling of uncertainty.
CausalWorld~\cite{ahmed2020causalworld} introduces a simulation benchmark for transfer learning in manipulation with physical attributes. However, it does not provide a principled causal reasoning framework for decision-making. Similarly, physics-based reasoning approaches~\cite{beetz2012ieee} model dynamics but lack the causal semantics required for probabilistic causal inference as enabled in our work.

Recent advances in probabilistic programming languages (PPLs), such as Pyro~\cite{Bingham2018Pyro}, enable flexible, generative causal modelling over arbitrary joint distributions. This supports structured, modular design for robot reasoning under uncertainty.
Our architecture builds on this by implementing a Pyro-based CBN, which can be extended to a structural causal model (SCM)~\cite{Pearl2009Causality} for future counterfactual inference~\cite{Cannizzaro2023CARDESPOT}.
Counterfactual reasoning has been shown to align with human causal judgements~\cite{Gerstenberg2022CounterfactualsVsHypotheticalsInHumanCausalJudgements} and is viewed as a critical component for building explainable and accountable robot systems~\cite{Hellstrom2021CausationInRobotics,Winfield2022EBB,Salvini2023AutonomousDecisionMakingSystems}. Our architecture lays the foundation for future work on counterfactual-based causal explanations for robot behaviour and decision outcomes.
%
\section{Background} \label{sec:background_knowledge}
Causality is the science of reasoning about cause and effect~\cite{Pearl2018Book}. In machine learning, causal models encode the relationships underlying a system’s data-generation process. Unlike purely statistical associations, causal models support intervention analysis and can correct for confounding bias caused by unobserved variables~\cite{Pearl2009Causality}. Failing to account for such biases can lead to \textbf{incorrect predictions} and consequently \textbf{unreliable action selection} or \textbf{explanation generation} in autonomous systems~\cite{Cannizzaro2023CARDESPOT, Bareinboim2015ConfoundedBandits, Ortega2021DelusionsInSequenceModelsForControl}.

Causal Bayesian networks (CBNs) are probabilistic graphical models that represent causal relationships via a directed acyclic graph (DAG)~\cite{Pearl2009Causality}. A causal DAG $G=\langle V, E \rangle$ consists of random variables $V$ and directed edges $E$ indicating conditional dependencies. Together with conditional probability distributions, the CBN defines a joint distribution over variables. Crucially, CBNs support interventional reasoning using $do(\cdot)$ calculus. An intervention on a variable $X$ alters the causal process, allowing us to infer how this affects another variable $Y$, as in $P(Y \mid do(X = x))$~\cite{Pearl2009Causality}.
\section{Causal Bayesian Reasoning Architecture} \label{sec:architecture}

We present \emph{\architectureacronym} (Fig.~\ref{fig:reasoning_framework_architecture_diagram}), a novel architecture for robot manipulation under uncertainty. For the first time, it combines CBN modelling, probabilistic programming, and online physics simulation to support generalisable and extensible decision-making across a wide range of robot morphologies, sensing modalities, and environments.

The architecture bridges low-level perception and control with high-level knowledge and reasoning, enabling the robot to pose questions about \textbf{what} it should believe given incomplete knowledge of the world (e.g., \emph{Should I expect the observed block tower to be stable?}), and \textbf{how} it should act to achieve its goals (e.g., \emph{Where should I place this block?}).

To illustrate its practical application, Fig.~\ref{fig:reasoning_framework_architecture_diagram} shows the integration of the architecture into a mobile robot system for the exemplar block stacking task (Sec.~\ref{sec:exemplar_block_stacking_task}--\ref{subsec:gazebo_robot_simulation}).

\subsection{Robot Sequential Decision-Making Causal Model}
At the core of \architectureacronym\ is a dynamic CBN that models the robot-world system across discrete time steps. This formulation captures the probabilistic dynamics governing the evolution of system state during robot interaction and enables both prediction and intervention-based inference.

We model the robot sequential decision-making process as a dynamic CBN due to its generality and flexibility. CBNs have been successfully applied to a wide range of agent-based decision-making, planning, and bandit problems, across both fully and partially observable domains and under stochastic or deterministic transitions~\cite{Cannizzaro2023CARDESPOT,Bareinboim2015ConfoundedBandits}. The CBN representation is also inherently compatible with many existing planning and reinforcement learning solvers.

Crucially, the model offers a \textbf{unified representation} for both sampling and inference. A single model definition supports generative data sampling (e.g., simulation rollouts) and probabilistic inference (e.g., conditioning on observations), enhancing modularity and reuse across tasks and domains.

\subsubsection{Discrete Time POMDP Model Abstraction}
We implement the CBN using a discrete-time partially observable Markov decision process (POMDP)~\cite{Smallwood1973}, a standard formalism for decision-making under uncertainty~\cite{Bai2015DESPOTAutonomousDriving,Budd2022,Cannizzaro2023CARDESPOT}. A POMDP is defined by the tuple $\langle \mathcal{S}, \mathcal{A}, T, \mathcal{Z}, O \rangle$, where $\mathcal{S}$ is the set of world states, $\mathcal{A}$ is the set of agent actions, $T(s, a, s')$ defines transition probabilities, $\mathcal{Z}$ is the set of observations, and $O(s', a, z)$ defines observation likelihoods.

The dynamic CBN uses time-indexed variables (e.g., $S_t$, $A_t$, $Z_t$) and models observation noise ($W_{Z_t}$) and actuation stochasticity ($W_{A_t}$), as shown in Fig.~\ref{fig:block_stacking_task_dag}. In fully observable domains, it reduces to a Markov decision process (MDP) by omitting $Z_t$ and making $A_{t+1}$ directly dependent on $S_t$.

Although we use a POMDP abstraction in our exemplar task, the model can represent any Python-computable probabilistic program. Inference variables must be defined using Pyro \emph{sample} statements, but this imposes no practical constraint. Pyro supports a large collection of common distributions (e.g., Gaussian, Categorical), hierarchical structures via plate notation, and also allows users to define custom distributions as needed. This enables the architecture to flexibly support a wide range of robot morphologies, actuation and sensing modalities, and domain-specific stochastic processes.

\subsubsection{Unlocking the Power of the Python Ecosystem}
We implement the robot decision-making model as a generative probabilistic program in Pyro, a probabilistic programming language (PPL) built on Python. 
Consequently, our PPL-based architecture can handle \textbf{modelling distributions as arbitrary general-purpose Python code}, and is not limited to closed-form analytical functions or differentiable programs as many other probabilistic methods are.
In addition to a flexible modelling capability, this enables integration with a range of scientific and robotics packages, allowing the construction of causal models using existing libraries.

In our exemplar task, the model leverages the PyBullet simulator during inference to evaluate candidate actions and task outcomes.
Beyond physics, Python provides access to scientific libraries for tasks such as signal modelling (e.g., for lidar, image, or wireless propagation), object classification, human pose estimation, and human cognitive models for human-robot interaction.
These components can be embedded directly into the causal model to support reasoning over perceptual, social, and contextual variables.

The architecture also supports pretrained models via Python-based ML libraries. These include large language models (LLMs) for inferring symbolic goals from instructions, and multi-modal models that produce latent variables (e.g., scene or intent embeddings) to condition simulation rollouts. Such components enhance causal reasoning in tasks requiring language, perception, or contextual understanding.

\subsection{Intervention-Based Causal Inference}
The modelling flexibility enabled by our PPL-based architecture also extends to inference: once the generative model is defined in Pyro, it can be paired with a range of inference algorithms to support prediction and decision-making over arbitrary-code models.
Representing the model in Pyro enables both exact and approximate inference methods, including sample-based (e.g., importance sampling~\cite{Kloek1978ImportanceSampling}) and gradient-based methods (e.g., stochastic variational inference, SVI~\cite{Wingate2013VariationalInference}).
Pyro also permits flexible composition of conditioning statements with interventions via the $do(\cdot)$ operator, allowing estimation of interventional transition posteriors conditioned on robot observations. 
These posteriors support predictive queries over current and future task states for decision-making.
This end-to-end approach enables tractable inference over causal models that incorporate rich, domain-specific simulators and perception modules.
\subsection{Software Interface for Hardware Integration}
\label{subsec:software_interface_for_hardware_integration}
To maximise the benefit to the robotics research community, our reasoning architecture includes software interfaces that expose the intervention-based causal inference functionality through both a pure Python API and a ROS action server API (Fig.~\ref{fig:reasoning_framework_architecture_diagram}).
In Section~\ref{sec:exemplar_block_stacking_task}, we demonstrate how the ROS interface integrates into a mobile robot system for the exemplar block stacking task.
%
\section{Exemplar Block Stacking Task}
\label{sec:exemplar_block_stacking_task}
\begin{figure}[t]
    \centering
    \begin{subfigure}{\robotdemosubfigwidth} 
        \centering
        \includegraphics[trim=7cm 0cm 13cm 5cm, clip, width=\columnwidth]{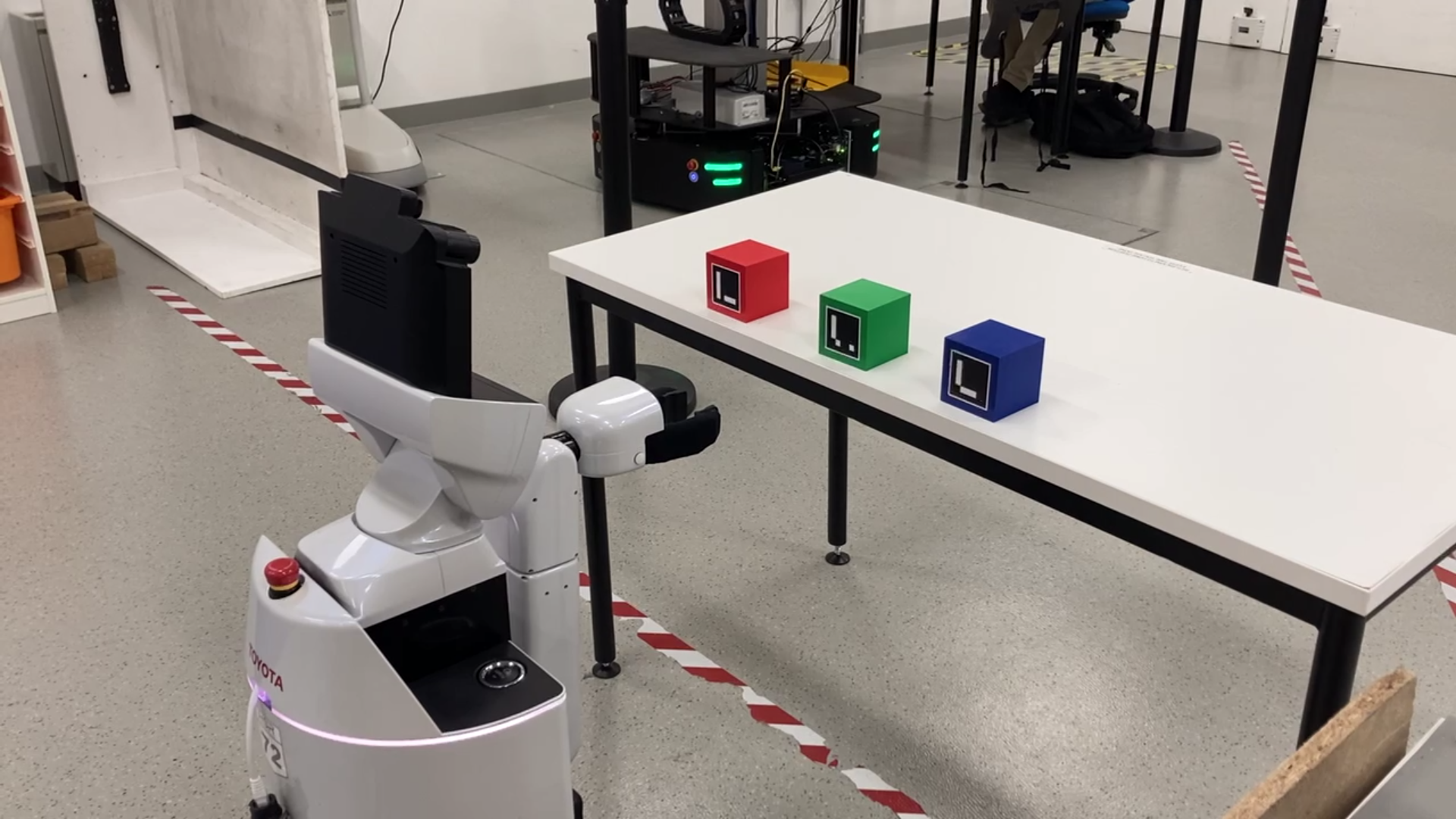}
        \vspace{-16pt}
        \caption{}
        \vspace{3pt}
    \end{subfigure}
    \begin{subfigure}{\robotdemosubfigwidth}
        \centering
        \includegraphics[trim=0cm 1.5cm 15cm 0cm, clip, width=\columnwidth]{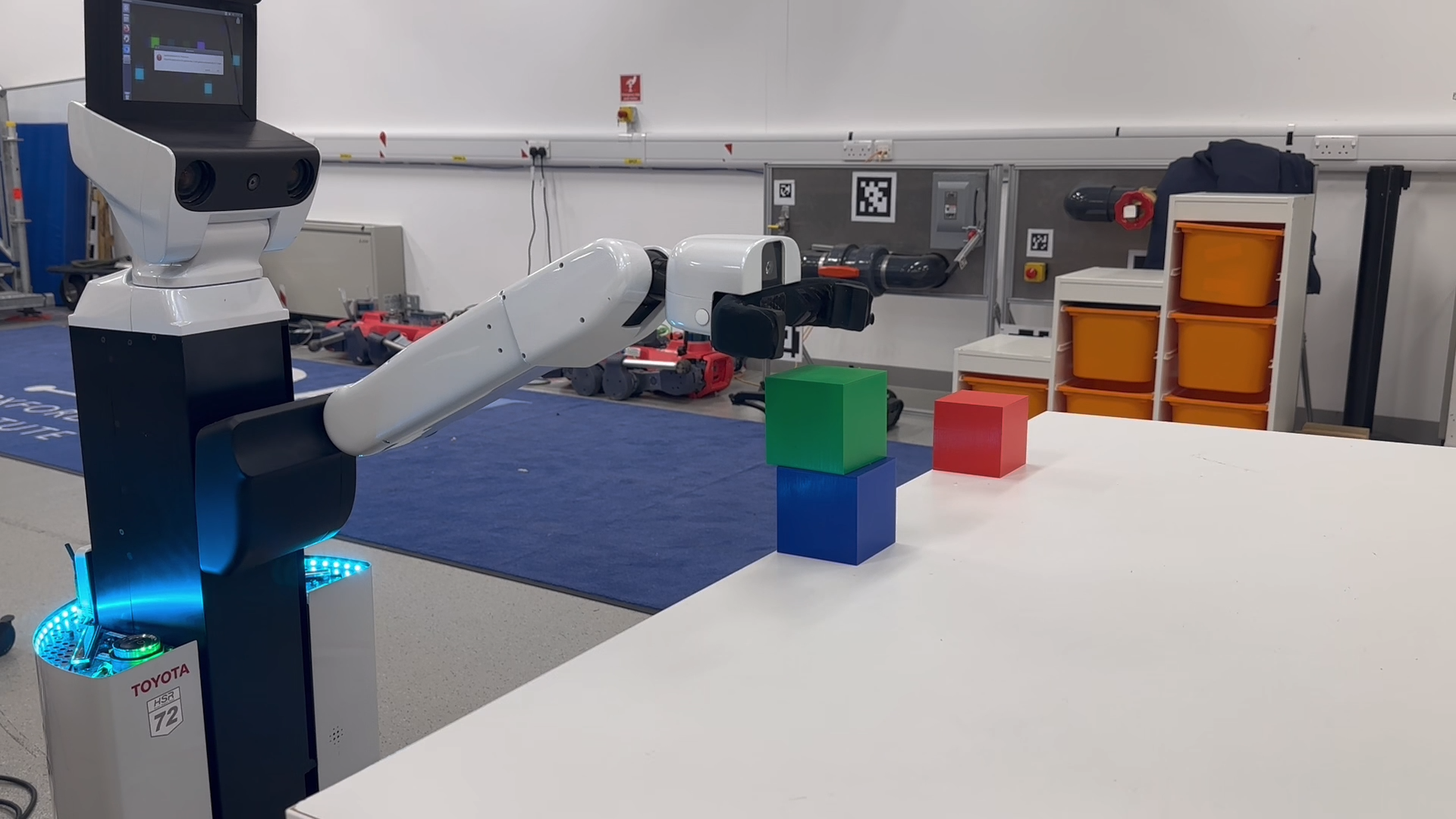}
        \vspace{-16pt}
        \caption{}
        \vspace{3pt}
    \end{subfigure}
    \begin{subfigure}{\robotdemosubfigwidth}
        \centering
        \includegraphics[trim=6cm 5cm 14cm 0cm, clip, width=\columnwidth]{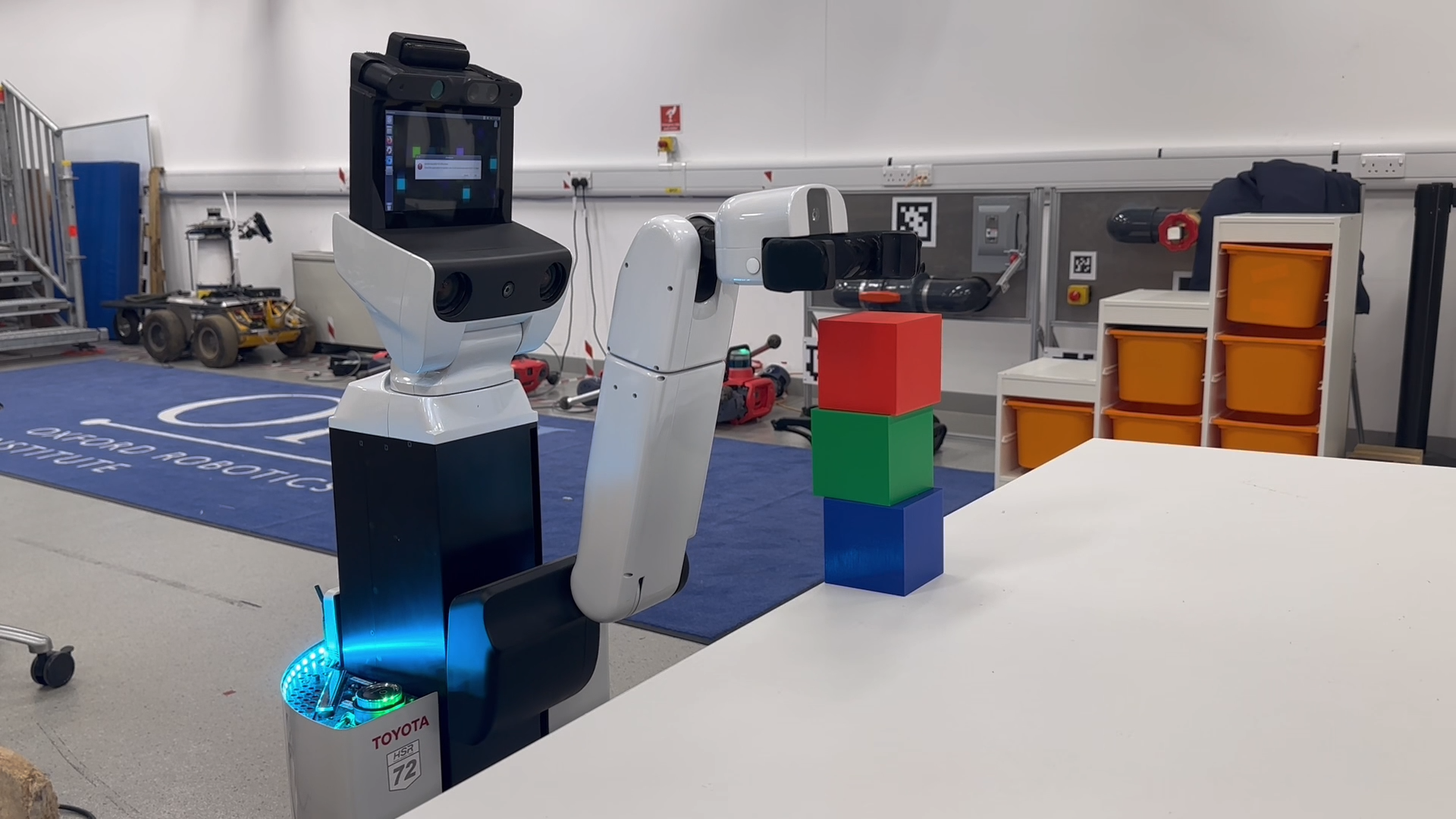}
        \vspace{-16pt}
        \caption{}
        \vspace{3pt}
    \end{subfigure}
    \begin{subfigure}{\robotdemosubfigwidth}
        \centering
        \includegraphics[trim=10cm 15.35cm 25cm 0cm, clip, width=\columnwidth]{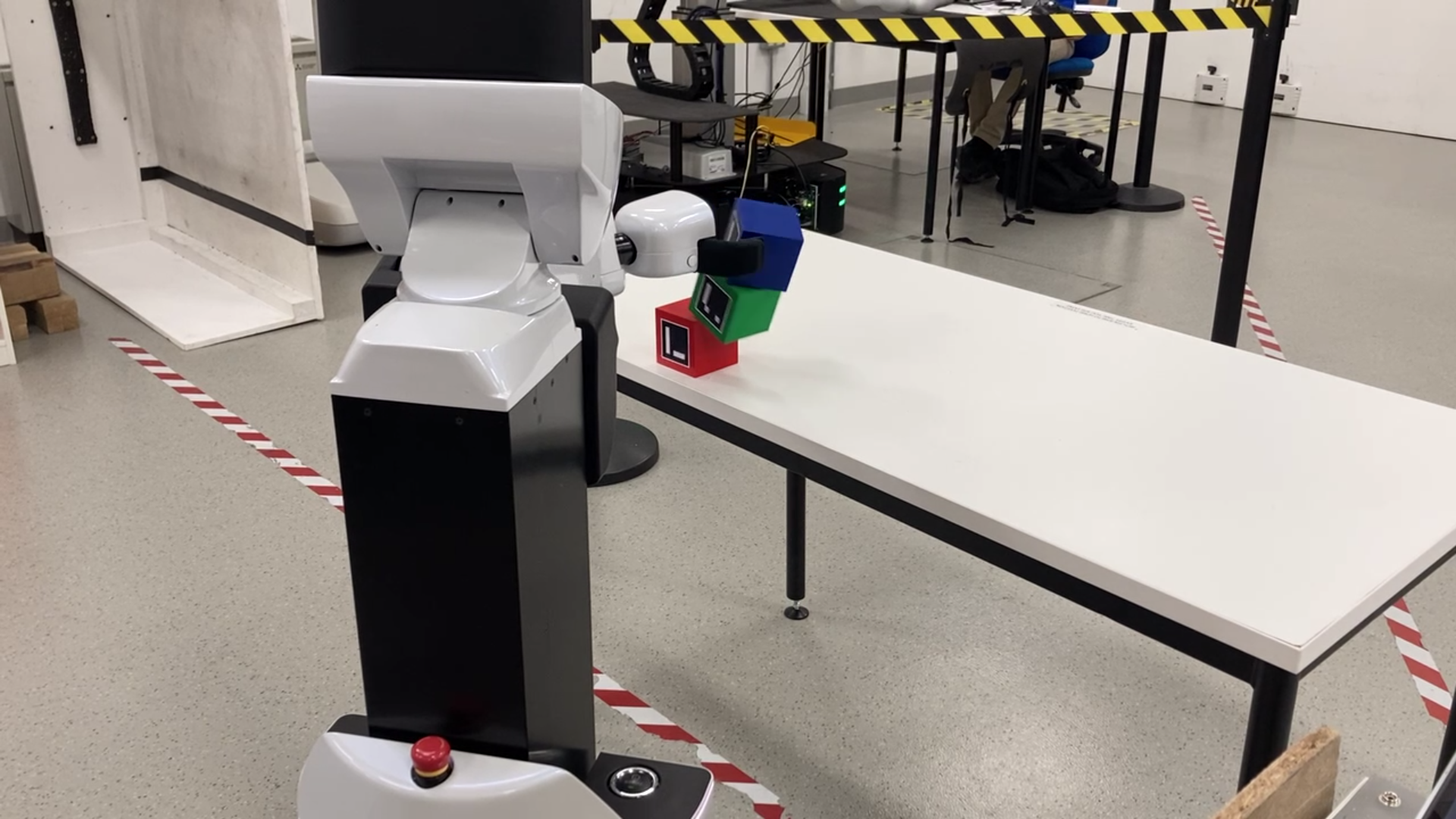}
        \vspace{-16pt}
        \caption{}
        \vspace{3pt}
    \end{subfigure}
    \vspace{-6mm}
    \caption{Real-world demonstration of the two-action block stacking task used to evaluate our causal reasoning architecture, showing the robot executing on physical hardware, progressing from: (a) the initial state, (b) an intermediate stable-tower state after the first action, to two possible outcomes: (c) a stable tower indicating success, or (d) an unstable tower indicating failure.
    }
    \label{fig:real_world_robot_block_stacking_task}
    \vspace{-4mm}
\end{figure}

\begin{figure}[t]
    \centering
    \includegraphics[trim={0 0.9cm 0 0},clip,width=0.9\columnwidth]{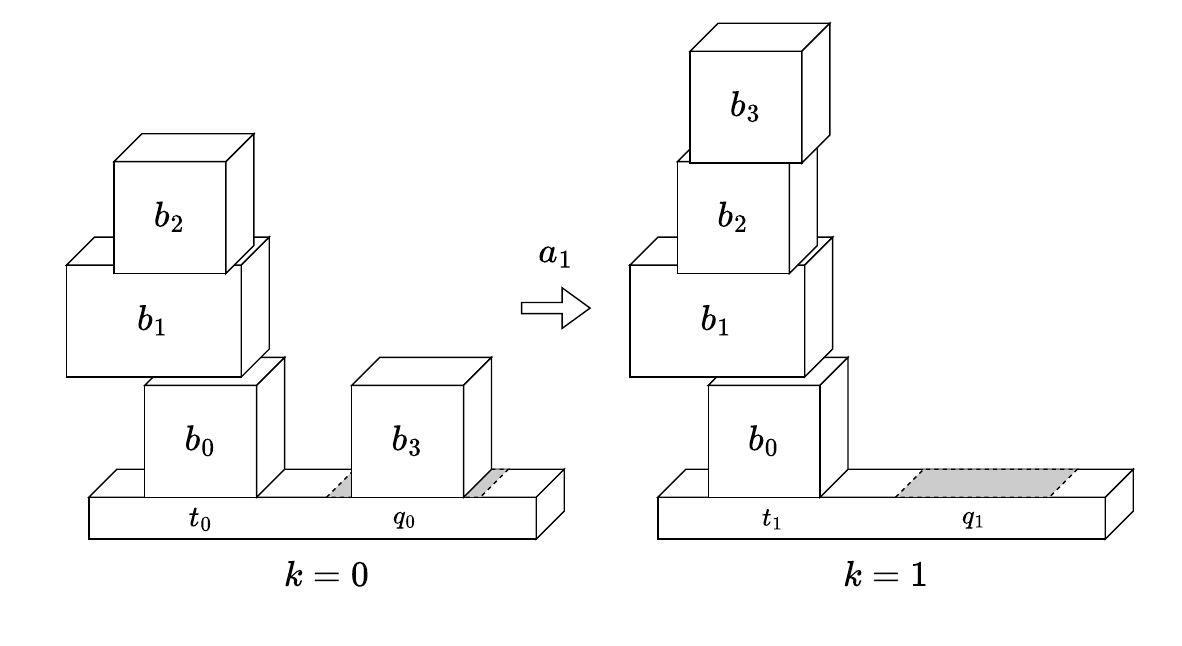}
    \caption{Schematic illustration of the exemplar block stacking task. At each time step $k$, the robot selects action $A_k$ to place block $b$ from queue $q_{k-1}$ onto previous tower $t_{k-1}$, resulting in the new state $(t_k, q_k)$. The goal is to incrementally construct a stable tower through sequential placements.
    } 
    \label{fig:block_stacking_problem_diagram}
    \vspace{-7mm}
\end{figure}

To demonstrate the practical application of our reasoning architecture, we apply it to an exemplar manipulation task: sequential block stacking (Fig.~\ref{fig:real_world_robot_block_stacking_task},~\ref{fig:block_stacking_problem_diagram}). The robot incrementally builds a tower from an initial configuration and a queue of blocks, using noisy sensor observations and stochastic placement actions.
The task is successful if the tower remains standing after the final placement and a failure if it topples at any point. A key challenge is managing uncertainty in sensing, state estimation, and control, all of which must be accounted for to ensure reliable execution.

As our focus is on a generalisable formulation for probabilistic prediction and action selection\ \textemdash\ rather than on planning methods and search\ \textemdash\ we restrict the task to single-column towers and frame it as a sequence of independent next-best action selection problems. This allows us to demonstrate the reasoning capabilities of the model without introducing additional planning complexity. Nevertheless, the model can be extended with a (PO)MDP planner to support trajectory-level optimisation when required~\cite{Cannizzaro2023CARDESPOT}.

\subsection{Problem Definition}
Formally, the robot's task is to construct a stable block tower by sequentially placing one block at each discrete time step $k = 1, \dots, K$, where $K$ is the total number of additional blocks to be placed. The task state at time $k$, denoted $s_k$, consists of the current tower configuration $t_k$ and the queue of remaining blocks $q_k$: $s_k = t_k \cup q_k$.

We formulate the greedy next-best placement problem as selecting an approximately optimal action $\hat{a}_k$ from the set of candidates $\mathcal{A}$. Each action $a = (x, y)$ corresponds to placing the next block at a position on top of the current tower. The goal is to select the action that maximises the probability that the resulting state $s_k$ is stable, conditioned on the previous (latent) state $s_{k-1}$ and its noisy observation $z_{k-1}$:
$\hat{a}_k = \underset{a \in \mathcal{A}}{\mathrm{argmax}} \left\{ P\left( \mathrm{IsStable}(s_k) \mid z_{k-1},\ a \right) \right\}$.
%
%
\subsection{Exemplar Task Decision-Making Causal Model}
\label{subsec:exemplar_task_decision_making_causal_model}
\subsubsection{Task Causal DAG}
Following the decision-making causal model formulation of our architecture (Section~\ref{sec:architecture}), we model the $K$-action block stacking task as a dynamic CBN. A dynamic causal DAG for a two-action ($K=2$) block stacking task is shown in Fig.~\ref{fig:block_stacking_task_dag}.
To ground the model in our task requirements, our exemplar causal model captures two main sources of uncertainty in the robot-task system: observation noise $W_{Z,k}$ and actuation noise $W_{A,k}$ at each time step $k$, which perturb the robot's environment observation $Z_k$ and execution of action $A_k$, respectively.
%
%
%
\subsubsection{Task State Representation}
The task state at time $k$, denoted $s_k$, consists of the current tower configuration $t_k$ and the queue of remaining blocks $q_k$: $s_k = t_k \cup q_k$. An example showing the initial state $s_0$ and resulting state $s_1$ after action $a_1$ is shown in Fig.~\ref{fig:block_stacking_problem_diagram}.
The tower state $t_k$ is represented as an ordered list of block references:
$t_k = [b_{n_i,k}]$ for $i = 0, \dots, I_k$,
where $n_i$ is the unique identifier of the block at position $i$ in the tower, and $I_k + 1$ is the number of blocks in the tower at time $k$.
Similarly, the queue state is:
$q_k = [b_{n_j,k}]$ for $j = 0, \dots, J_k$,
where $n_j$ is the unique identifier of the block at position $j$ in the queue, and $J_k + 1$ is the number of remaining blocks.
Each block state $b_{n,k}$ encodes the physical attributes of block $n$ at time $k$, including its 6-DOF pose (position and orientation), 3D dimensions, and mass.
\subsubsection{Action Representation}
Each action $a \in \mathcal{A}$ is defined as a 2D placement coordinate $a = (x, y)$, specifying where the next block is to be placed on top of the current tower.
%
%
%
\subsubsection{Transition Function Representation}
\begin{figure}[t]
    \centering
    \includegraphics[trim=0cm 0cm -1cm 0cm, clip,width=0.9\columnwidth]{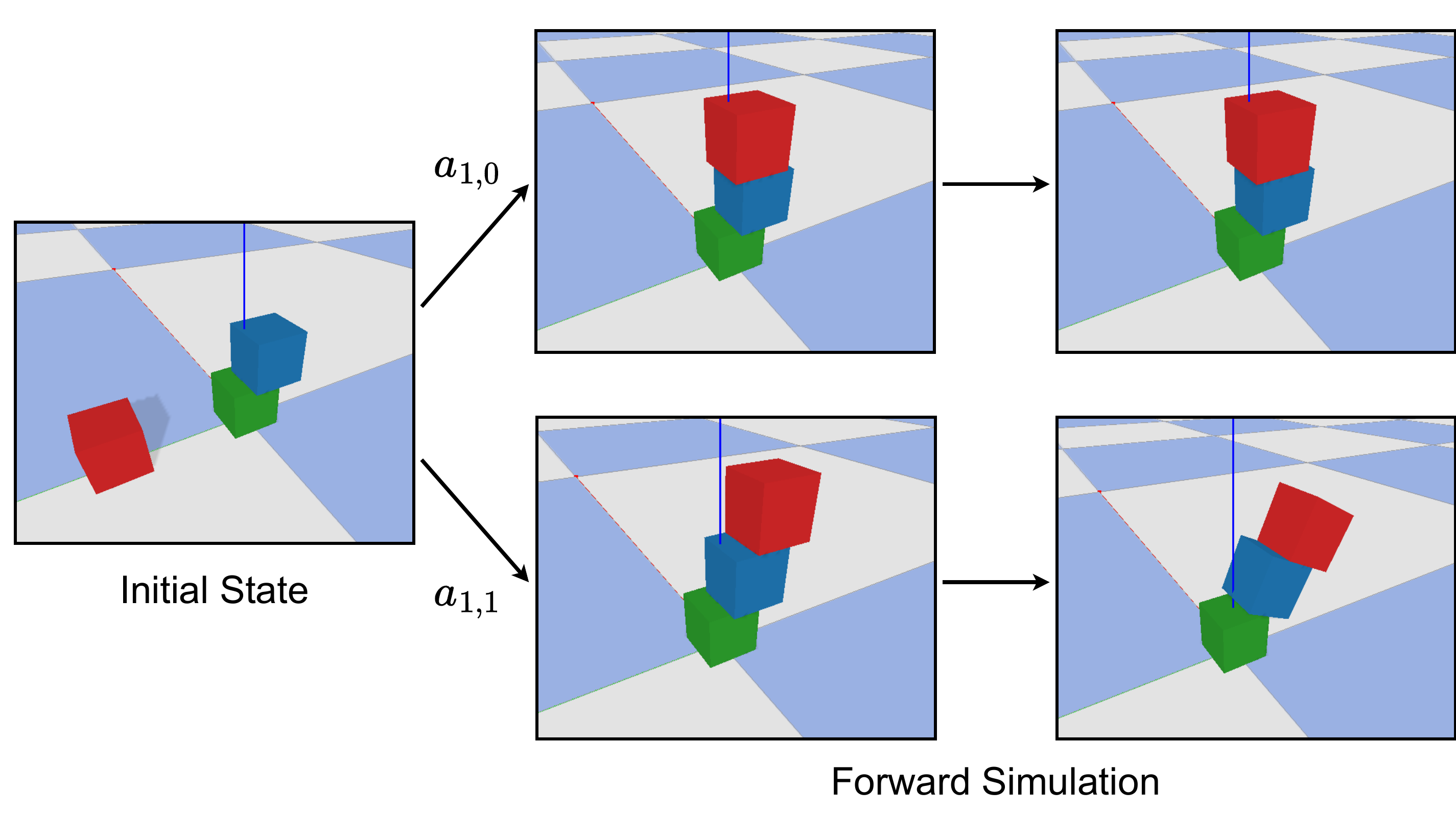}
    \caption{Physics-based PyBullet simulation used in the architecture decision-making causal model to predict successor state tower stability probability.} \label{fig:pybullet_block_tower_simulation}
    \vspace{-7mm}
\end{figure}

We model the state transition function $T$ following the standard POMDP formulation~\cite{Smallwood1973}, adopting a causal perspective by treating the agent’s action as an intervention using the $do(\cdot)$ operator. 
This allows $A_k$ to be interpreted as a direct manipulation of the system, marginalising over execution noise $W_{A,k}$:
$T(S_{k-1}, A_k, S_k) = P\big(S_k \mid S_{k-1}, do(A_k = a_k)\big)$.
Uncertainty in observation and action outcome is introduced by sampling error terms $W_{Z,k}$ and $W_{A,k}$ from their respective noise distributions. 
To sample the successor state $s_k$ from the transition distribution $T$ during CBN-based model sampling and inference, we use the PyBullet 3D physics simulator~\cite{coumans2021} to simulate the 6-DOF rigid body dynamics of the tower (Fig.~\ref{fig:pybullet_block_tower_simulation}), conditioned on the agent’s noisy observation $Z_{k-1}$ of the previous state. 
These samples are propagated into the initial PyBullet simulation state, perturbing the placement pose of the new block away from the robot’s intended action $(x, y)$. 
The simulator then performs a deterministic forward simulation of the physical system.
The resulting tower configuration is used to determine the successor state.
%
%
\subsubsection{Noise \& Error Modelling}
\label{subsubsec:block_stacking_noise_modelling}
We model both observation and manipulation errors as zero-mean, isotropic Gaussian noise in 3D. Specifically, observation noise and action execution error are represented as independent random vectors with components drawn from $\mathcal{N}(0, \sigma_Z^2)$ and $\mathcal{N}(0, \sigma_A^2)$, respectively.
The standard deviations $\sigma_Z$ and $\sigma_A$ represent uncertainty in the robot’s perception and control, due to sensor noise, estimation errors, and actuation imprecision. 
Noise terms are assumed independent across axes and time.
%
%
%
\subsection{Evaluation Tasks}
We evaluate our architecture on two tasks: (1) tower stability prediction, and (2) greedy next-best action selection.
\subsubsection{Task 1: Tower Stability Prediction}
\label{subsec:tower_stability_prediction}
We frame tower stability prediction as a binary classification problem: estimating the probability that a perceived tower configuration remains stable in the successor state. 
We use our causal model to estimate the query $\Phi_{stable,k}$, the probability that the unobserved true tower state $S_{k+1}$ remains stable, conditioned on the robot’s noisy observation $z_k$: $\Phi_{stable,k} = P(\mathrm{IsStable}(S_{k+1}) = \mathrm{True} \mid z_k)$. We apply a threshold value $\tau_{stable,Z}$ to convert the probability into a binary decision.

The threshold $\tau_{stable,Z}$ acts as a tuning parameter representing tolerance to uncertainty from perception noise. Systems with higher measurement error may require a more conservative threshold to reduce false positives. In safety-critical tasks such as tower construction or fault detection, minimising false positives is crucial. The cost of misclassifying an unstable tower may far exceed that of a missed opportunity, e.g., a block falling onto a nearby human.
\subsubsection{Task 2: Greedy Next-Best Action Selection}
\label{subsec:next_best_action_selection_method}
We pose greedy next-best action selection as an optimisation problem: finding the action that maximises the predicted probability of the tower remaining stable after block placement.

We begin by generating candidate actions $a = (x, y)$ via uniform sampling from an $L \times W$ grid over the top face of the current block. 
For each candidate, we compute the intervention-based inference query $\Phi_{stable,k,a}$ using our CBN model, estimating the probability that the tower remains stable in the successor state $S_k$, conditioned on the robot’s observation $z_{k-1}$ and the intervention $do(A_k = a_k)$.

We define the filtered action set $\mathcal{A}_{\tau}$ as those candidates whose predicted stability probability exceeds the minimum threshold $\tau_{stable,A}$. The most stable action is then selected:
{\small$
a^*_k = \underset{a_k \in \mathcal{A}_{\tau}}{\mathrm{argmax}}\ P\big(\mathrm{IsStable}(S_k) = \mathrm{True} \mid do(A_k = a_k),\ z_{k-1}\big)
$.}

We define the stable set $\mathcal{A}_{stable}$ as actions from 
$\mathcal{A}_\tau$ that are within a probability margin $\tau_{cluster}$ of $p_{a^*_{k}}$:
\mbox{{\small
$
\mathcal{A}_{stable} = \{a \in \mathcal{A}_\tau : |p_{a^*} - p_a| \leq \tau_{cluster} \}
$.}}
%
We assume the stable set $\mathcal{A}_{stable}$ forms a convex hull approximating the region of stable placements. We compute the geometric mean of the $(x, y)$ positions of actions in this set and select the centroid as the next-best action $a_k$.
Other decision rules, such as conditional value-at-risk (CVaR), can also be implemented within our architecture using the same inferred posteriors.

The post-placement stability threshold $\tau_{stable,A}$ defines the minimum acceptable probability for action success. The cluster threshold $\tau_{cluster}$ controls how close an action must be to $a^*_k$ to be included in the geometric mean. Like $\tau_{stable,Z}$, both should be tuned based on the application’s risk profile.
%
\section{Experimentation} 
\label{sec:experimentation}
%
\subsection{High-Fidelity Gazebo Robot Simulation Evaluation}
\label{subsec:gazebo_robot_simulation}
We evaluate our architecture on the exemplar block stacking task using a simulated Toyota Human Support Robot (HSR)
in Gazebo.
The setup uses three 7.5~cm cube blocks matching those in the real-world demonstration (Fig.~\ref{fig:real_world_robot_block_stacking_task}). 
To simulate realistic block position estimation, we attach ArUco markers to each block and generate 6-DOF pose observations from simulated RGB-D data using the OpenCV ArUco module. Pick-and-place actions are executed via motion plans generated by ROS MoveIt!.

\subsection{Task 1: Tower Stability Prediction}
\subsubsection{Block Pose Estimation Error Characterisation}
To characterise the error in the robot’s block position estimation, we generate a dataset of 250 randomly sampled 3-block tower configurations in Gazebo simulation. For each configuration, we record the 3D positions of all blocks as observed by the robot, along with their ground-truth positions in the simulation environment. 
The estimation error is quantified by computing the standard deviation of the position error along each axis: $\sigma_{Z,x}$, $\sigma_{Z,y}$, and $\sigma_{Z,z}$. Empirically, we find that using the average of these values provides a sufficiently accurate approximation for a shared standard deviation $\sigma_Z$, applied across all axes in the zero-mean, isotropic 3D Gaussian noise model (see Sec.~\ref{subsubsec:block_stacking_noise_modelling}).
\subsubsection{Tower Stability Classification Accuracy}
We generate a dataset of 1000 randomly sampled 3-block tower configurations in simulation to evaluate model classification accuracy.
Ground-truth stability labels are assigned by forward-simulating each  configuration in Gazebo and observing whether the tower remains standing.
We use the Importance Sampling inference method in Pyro, drawing 50 samples per query, to estimate tower stability based on our exemplar task causal model (see Sec.~\ref{subsec:exemplar_task_decision_making_causal_model}) and the estimated value of $\sigma_Z$. 
Binary classifications are produced by thresholding the predicted stability probability.
To assess performance, we compute standard classification metrics ($\text{F}_1$ score, AUC, precision, and recall) across a range of thresholds.
\subsection{Task 2: Greedy Next-Best Action Selection}
\subsubsection{Block Placement Error Characterisation}
To characterise block placement error using the robot’s manipulation subsystem, we generate a dataset of 25 randomly sampled 2-block tower configurations in Gazebo. Each defines the initial tower state for a placement trial.
For each configuration, the robot performs 10 independent attempts to place an additional block at a predefined target position, resulting in 250 placement experiments. After each attempt, the placement error (i.e., deviation from the intended location) is recorded.
Placement error is quantified by computing the standard deviation along each axis: $\sigma_{A,x}$, $\sigma_{A,y}$, and $\sigma_{A,z}$. We find that the average of these values provides a sufficiently accurate approximation for a shared standard deviation $\sigma_A$, applied across all axes in the zero-mean, isotropic 3D Gaussian model representing manipulation uncertainty.
\subsubsection{Greedy Next-Best Action Selection Performance}
To evaluate system performance on the greedy next-best action selection task in simulation, we generate a test dataset of 50 randomly sampled initial 2-block tower configurations.
For each configuration, we use our action selection method and causal model to choose a placement position for a third block (i.e., a single-action task with $K = 1$). Candidate actions are sampled from a uniform $5 \times 5$ grid over the top face of the current block. Inference is done using Pyro's Importance Sampling with 50 samples per query.
Each selected action is executed in 10 independent trials, yielding 500 total experiments. These repetitions are used to compute empirical success rates.
We set the stability threshold $\tau_{stable,A} = 0.8$ to ensure high-confidence actions. The clustering threshold is $\tau_{cluster} = 0.2$, admitting all candidates above $\tau_{stable,A}$ into the stable set.
This looser clustering highlights the benefit of centroid-based action selection.
%
%
\subsubsection{Naive Baseline Action-Selection Method} 
We compare our method against a baseline that follows a naive, heuristic policy: always placing the next block at the centre of the current top block.
This simple strategy can often produce stable towers under ideal conditions. However, it lacks any understanding of block physics, system dynamics, or task uncertainty, and is not expected to make robust placement decisions under realistic noise and variability.
%
\section{Results \& Discussion} \label{sec:results_and_discussion}
%
\subsection{Task 1: Tower Stability Prediction}
\subsubsection{Block Pose Estimation Error Characterisation}
Results from the block position error characterisation are shown in Table~\ref{table:position_measurement_and_placement_error_parameters}. 
The data reveal that our initial assumption, that the random error term $W_Z$ is independent of the robot’s state, does not strictly hold.
In particular, measurement error along the X-axis, approximately aligned with the camera’s depth axis , exhibits a non-zero mean and substantially larger standard deviation than the Y- and Z-axes.
Despite this epistemic modelling error, it does not significantly degrade model performance. 
We therefore retain the simplifying assumption of zero-mean, state-independent noise. 
The empirical error distributions (not shown due to space constraints) are approximately symmetric and bell-shaped, supporting the use of a Gaussian approximation.
We take the average as an accurate estimate for a shared standard deviation $\sigma_Z = 0.469$\,cm, applied across all axes in our 3D Gaussian noise model.
%
\begin{table}[t]
    \centering
    \renewcommand{\arraystretch}{1.2}
    \caption{%
    Gazebo-based characterisation of block placement and measurement errors. Values show position error standard deviation (cm) along each axis. Model parameters (in-text) are computed as the mean across axes.
    }
    \begin{adjustbox}{width=1.0\columnwidth}
    \begin{tabular}{lcccc}
        \toprule
        \textbf{Error Type} & \textbf{X-axis} & \textbf{Y-axis} & \textbf{Z-axis} & \textbf{Avg. ($\sigma$)} \\
        \midrule
        Measurement ($\sigma_Z$) & 0.906 & 0.216 & 0.284 & 0.469 \\
        Placement ($\sigma_A$)   & 1.790 & 2.770 & 0.146 & 1.570 \\
        \bottomrule
    \end{tabular}
    \end{adjustbox}
    \label{table:position_measurement_and_placement_error_parameters}
\end{table}
%
%
\subsubsection{Tower Stability Classification Accuracy}
\begin{figure}[t]
    \centering
    \includegraphics[width=1.0\columnwidth]{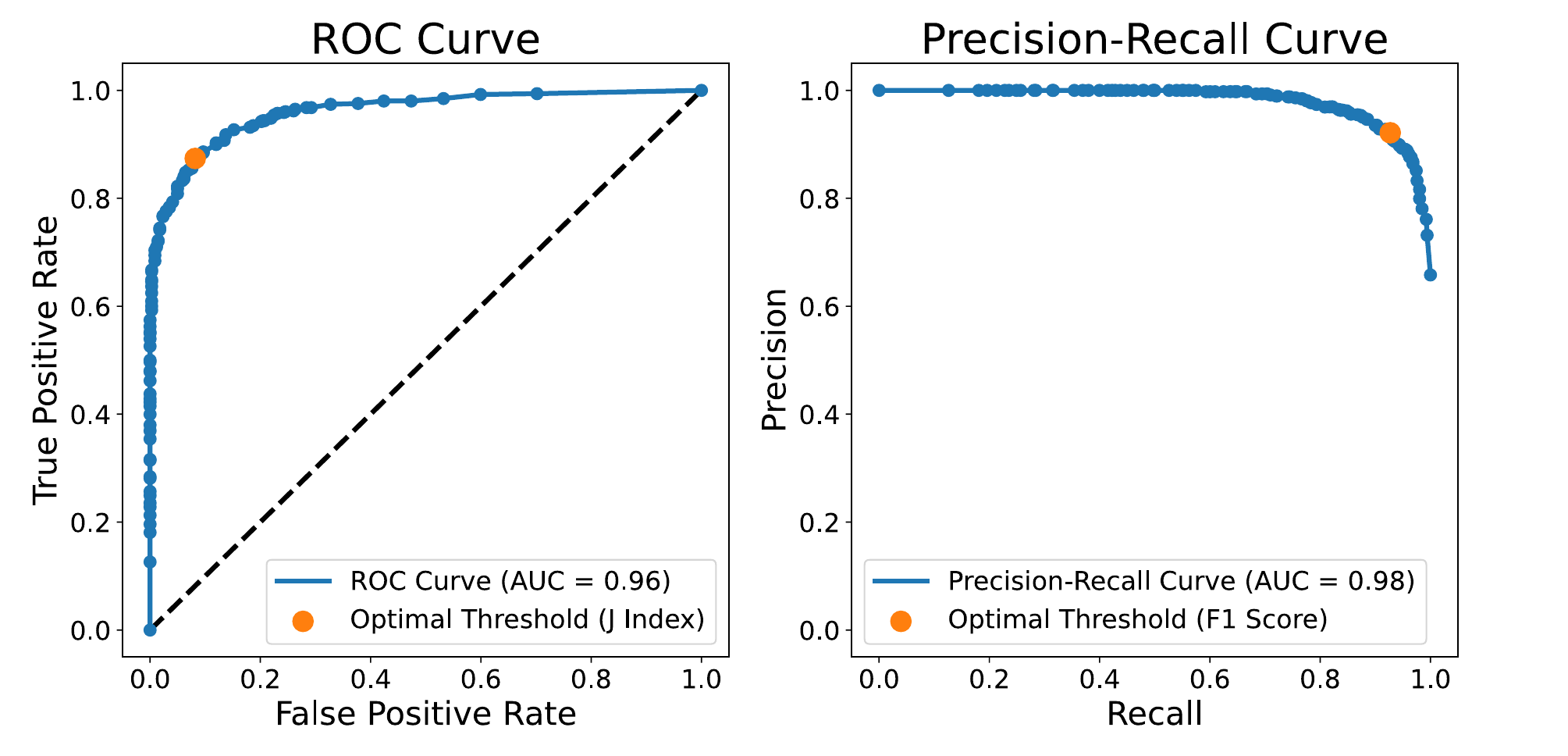}
    \caption{Receiver operating characteristic (ROC) and precision-recall (PR) curves for tower stability classification based on model predictions. The curves demonstrate strong overall binary classification performance, with the optimal classification threshold selected using Youden's Index \mbox{($\tau_{stable,Z} = 0.4$),} corresponding to a precision of 0.955 and recall of 0.868.}
    \label{fig:roc_and_precision_recall_curves}
\end{figure}

Model performance on the test set is shown in Table~\ref{table:prediction_results} and Fig.~\ref{fig:roc_and_precision_recall_curves}.
The best stability threshold was identified as 0.40 based on Youden’s index~\cite{youden1950index} from the ROC curve, corresponding to a precision of \precision\ and recall of \recall. Motivated by safety-critical considerations discussed in Section~\ref{subsec:tower_stability_prediction}, this threshold was selected for its lower false positive rate compared to the $\text{F}_1$-optimal threshold from the precision-recall curve.
These values are close to ideal classification performance and demonstrate the strong predictive ability of our model.
%
\begin{table}[t]
\centering
\renewcommand{\arraystretch}{1.2}
\caption{The model achieves almost ideal classification performance.}
\begin{tabular}{|c|c|c|c|c|c|}
\hline
\textbf{Metric}               & \begin{tabular}[c]{@{}c@{}}Prediction\\ Accuracy\end{tabular}    & \begin{tabular}[c]{@{}c@{}}F1\\ Score\end{tabular} & Precision                 & Recall                 & \begin{tabular}[c]{@{}c@{}}AUC\\ Score\end{tabular} \\ \hline
\textbf{\ \ Score} $(\uparrow)$\ & \textbf{ \predacc} & \fscore                             & \precision & \recall & \textbf{\aucscore}  \\ \hline
\end{tabular}
\label{table:prediction_results}
\vspace{-4mm}
\end{table}
%
%
%
\subsection{Task 2: Greedy Next-Best Action Selection}
\subsubsection{Quantitative Results}
\textbf{Block Placement Error Characterisation} \textemdash\  
Results are shown in Table~\ref{table:position_measurement_and_placement_error_parameters}.
As with measurement error, placement error along the Y- and X-axes exhibits substantially larger standard deviations than the Z-axis. The largest error is along the Y-axis, though this directional bias is not accounted for in the noise model.
Despite this anisotropy, the simplifying assumption of zero-mean, isotropic noise does not significantly degrade downstream performance. We therefore retain this assumption and take the mean of the axis-specific values to define a shared standard deviation of $\sigma_A = 1.57$\,cm to represent manipulation noise in our Gaussian error model.

\textbf{Simulated Robot Task Evaluation} \textemdash\  
Results of the next-best action selection evaluation using the simulated robot are shown in Table~\ref{tab:next_best_action_sim_results}. Our causal reasoning architecture achieves a \textbf{19.8 percentage point} increase in task success rate over the baseline. In the ideal case without manipulation error, the improvement reaches \textbf{30 percentage points}.

\begin{figure}[t]
    \centering
    \includegraphics[trim=0.0cm 2.4cm 0cm 0cm, clip, width=0.97\columnwidth]{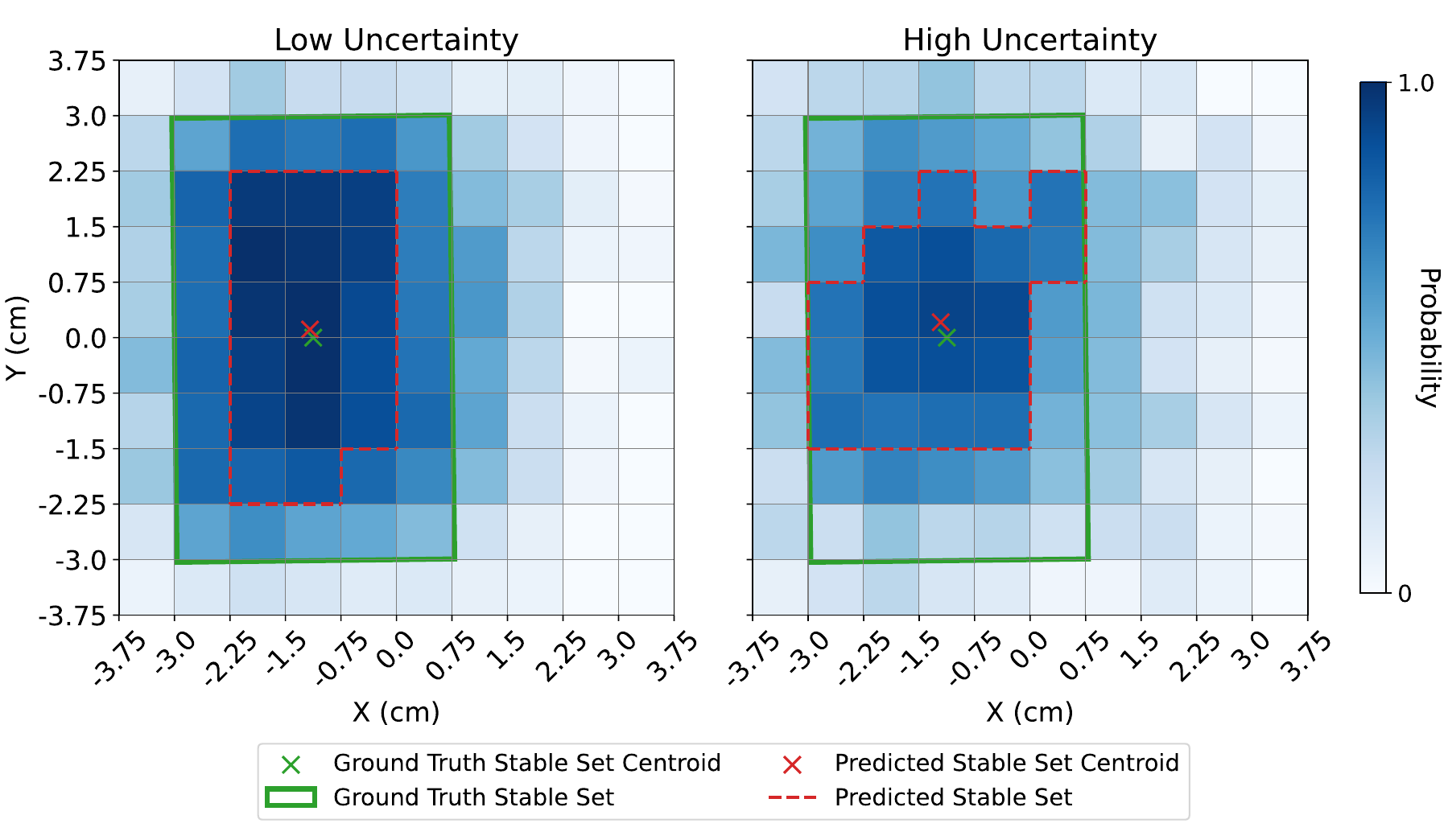}
    \caption{Predicted tower stability probabilities for candidate block placements under low (left) and high (right) system uncertainty, given the same initial tower. Predicted stable sets and centroids (robot’s actions) are shown in red; ground-truth stable regions and centroids in green. Higher uncertainty shrinks and centralises the predicted stable set due to edge risk and isotropic 3D Gaussian noise on the robot’s belief.}
    \label{fig:candidate_action_probability_heatmap}
    \vspace{-6mm}
\end{figure}

Fig.~\ref{fig:candidate_action_probability_heatmap} compares predicted stability probabilities for candidate placement positions during decision-making. The figure illustrates how our architecture accounts for both perception and manipulation uncertainty under low and high noise conditions, highlighting its robustness to real-world variability.
%
\begin{table}[t]
    \centering
    \caption{Performance on the block stacking task in simulation. Our architecture outperforms the naive baseline (\textbf{$\uparrow$19.8\%}) under realistic noise, and achieves \textbf{100\%} success without manipulation error (\textbf{$\uparrow$30\%}), highlighting the benefit of reasoning about physical stability and action uncertainty.}
    \renewcommand{\arraystretch}{1.2}
    \begin{adjustbox}{width=1.0\columnwidth}
    \begin{tabular}{
        >{\raggedright\arraybackslash}m{0.55\columnwidth} 
        >{\centering\arraybackslash}m{0.15\columnwidth} 
        >{\centering\arraybackslash}m{0.15\columnwidth} 
        >{\centering\arraybackslash}m{0.15\columnwidth}
    }
        \toprule
        \shortstack{\textbf{Action-Selection Method}} 
        & \shortstack{\textbf{Task}\\\textbf{Successes}} 
        & \shortstack{\textbf{Task}\\\textbf{Failures}} 
        & \shortstack{\textbf{Success}\\\textbf{Rate}} \\
        \midrule
        Baseline & 372 & 128 & 74.4\% \\
        \architectureacronym{} \textbf{(Ours)} & 471 & 29 & 94.2\% \\
        \midrule
        Baseline (No Manip. Noise) & 35 & 15 & 70.0\% \\
        \architectureacronym{} (\textbf{Ours}, No Manip. Noise) & 50 & 0 & \textbf{100\%} \\
        \bottomrule
    \end{tabular}
    \end{adjustbox}
    \label{tab:next_best_action_sim_results}
    \vspace{-10pt}
\end{table}
%
\subsubsection{Discussion}
The improved performance of our architecture over the baseline likely stems from its ability to explicitly model rigid-body physics and account for uncertainties in perception and manipulation.
By integrating the PyBullet 3D physics simulator into the causal model, our method simulates dynamics with high accuracy, enabling realistic predictions of candidate block placements. In contrast, the baseline uses a simple heuristic that ignores both dynamics and uncertainty.
Thus, our architecture generalises more effectively across varied initial tower configurations. 
By leveraging mental simulation for action selection, it remains robust even in challenging cases, e.g., when intermediate blocks are placed off-centre.
This robustness is further demonstrated by the \textbf{100\% success rate} in the ideal setting without manipulation error, highlighting the advantage of our method when precise action selection is critical.
\subsection{Architecture Scalability \& Limitations}
\label{sec:scalability_and_limitations}
Our architecture is designed to be modular and extensible, enabling application to a wide range of manipulation scenarios.
In this section, we explore its scalability and discuss limitations when applied to more complex tasks.
\subsubsection{Action Spaces}
The proposed dynamic CBN-based decision-making causal model can be extended to support richer action representations. 
In principle, additional action variables, such as diverse block geometries, frictional properties, or non-canonical placement orientations, can be incorporated by expanding the causal graph.
Pyro’s plate notation and conditional independence enable efficient factorisation of the joint distribution, improving scalability over flat representations.
Our model also supports multi-tower or interacting structure scenarios by modularising local subgraphs and using a physics-based simulator, such as PyBullet (as in our exemplar task), to capture relevant physical interactions.
\subsubsection{Sequential Decision-Making}
While the present work focuses on greedy, single-step reasoning, the framework is readily extensible to sequential settings such as MDPs and POMDPs. In these cases, the CBN-based probabilistic reasoning layer can inform higher-level control modules by modelling stochastic transitions under interventions.
Our formulation is compatible with CAR-DESPOT~\cite{Cannizzaro2023CARDESPOT}, a causal-aware POMDP planner that complements our approach. This enables hybrid architectures in which CBNs provide semantically grounded transition models, while planners coordinate long-horizon action sequences.
Such a composition retains the interpretability and structure-awareness of our framework, while enabling principled reasoning over policies and belief updates under partial observability.

\subsubsection{Reward Functions and Statistical Objectives}
The architecture supports flexible objectives beyond expectations over Gaussian rewards.
Risk-aware metrics such as CVaR, percentile thresholds, or multi-objective criteria can be incorporated without changes to the causal graph.
For example, a practitioner may wish to optimise for both task success and robustness to disturbances such as surface vibrations or wind. 
Probabilistic programming enables sampling-based estimation under arbitrary distributions, allowing non-Gaussian, asymmetric, or heavy-tailed beliefs.
This flexibility supports diverse task definitions, robot morphologies, sensing modalities, and deployment environments.

\subsubsection{Computational Limitations and Complexity}
A key limitation lies in the computational cost of inference when scaling to large causal graphs with many variables and dependencies. While exact inference is intractable in such settings, our current implementation uses importance sampling to approximate posterior distributions over latent variables. 
More scalable methods, such as stochastic variational inference (SVI), could be adopted in future work\ \textemdash\ for example, using Pyro’s plate notation to enable parallelism across samples.

Headless PyBullet simulation runs faster than real-time in our task and is not a major bottleneck. However, simulation cost may increase in more complex scenes, especially those involving many rigid bodies or advanced dynamics such as soft-body deformation or fluid interactions. 
Techniques such as simulation caching, surrogate models, or amortised inference offer promising paths to reduce computational load.

\subsubsection{Adaptability and Domain Transfer}
Although causal models are interpretable and structured, deploying them in new environments may require domain-specific assumptions or structure learning, introducing overhead compared to fully data-driven methods.
However, our framework is designed for modularity and reuse: the causal graph, inference engine, and decision-making components are loosely coupled and can be adapted independently. 
Causal models are also well suited to domain transfer, as they aim to capture fundamental causal mechanisms that govern system behaviour and are expected to remain invariant across environments.
This composability supports flexible transfer by reusing learned causal templates, swapping simulators, or integrating new inference targets without modifying the overall architecture.
%
\section{Real-World Robot Demonstration} 
\label{sec:robot_demonstration}
We deploy our architecture on a real-world HSR performing the exemplar block stacking task (Fig.~\ref{fig:real_world_robot_block_stacking_task}). 
The physical setup mirrors simulation, with one key variation: simulation trials begin with a two-block tower, while real-world trials start from a single block and place two sequentially, to demonstrate robustness to cumulative action-outcome uncertainty.
No parameters were re-tuned. 
The same perception and manipulation stack was used in both settings, and physical blocks matched simulation dimensions.

The system succeeded in \tasksuccessesrealrobot\ out of \taskattemptserealrobot\ trials,
demonstrating strong sim2real generalisation and suitability for deployment without re-tuning or retraining.
While initial configurations differ slightly, the core task remains unchanged: reasoning under uncertainty about tower stability and action consequences. 
A possible source of variation is that, in simulation, the robot places a third block onto randomly generated two-block towers, which may be less stable than the deliberately constructed intermediate two-block towers in real-world trials.
Nevertheless, our model explicitly accounts for perception and actuation noise, enabling robust performance across sequential placements despite cumulative uncertainty.
%
%
\section{Conclusion} 
\label{sec:conclusion}
We presented \architectureacronym, a novel causal Bayesian reasoning architecture that combines a decision-making causal model, probabilistic programming, and data-driven components to support robust manipulation under uncertainty. 
Our architecture enables predictive reasoning and greedy action selection in manipulation tasks through a structured, probabilistic model of physical interactions.
We validated our method in simulation, achieving high-accuracy prediction of manipulation outcomes and strong task performance in a challenging block stacking domain, and demonstrated sim2real generalisation by deploying the architecture on a domestic service robot, without retraining or parameter tuning.
This work contributes a generalised probabilistic reasoning architecture and opens new opportunities for causal and counterfactual reasoning in trustworthy autonomous systems.
%

\bibliographystyle{ieeetr}
\bibliography{IEEEabrv,references_cleaned_no_doi}

\end{document}